\def\year{2021}\relax
\documentclass[letterpaper]{article} % DO NOT CHANGE THIS
\usepackage{aaai21}  % DO NOT CHANGE THIS
\usepackage{times}  % DO NOT CHANGE THIS
\usepackage{helvet} % DO NOT CHANGE THIS
\usepackage{courier}  % DO NOT CHANGE THIS
\usepackage[hyphens]{url}  % DO NOT CHANGE THIS
\usepackage{graphicx} % DO NOT CHANGE THIS
\usepackage{graphicx}  % DO NOT CHANGE THIS
\usepackage{natbib}  % DO NOT CHANGE THIS AND DO NOT ADD ANY OPTIONS TO IT
\usepackage{caption} % DO NOT CHANGE THIS AND DO NOT ADD ANY OPTIONS TO IT
\usepackage[ruled,vlined]{algorithm2e}
\usepackage[utf8]{inputenc} % allow utf-8 input
\usepackage{url}            % simple URL typesetting
\usepackage{booktabs}       % professional-quality tables
\usepackage{amsfonts}       % blackboard math symbols
\usepackage{nicefrac}       % compact symbols for 1/2, etc.
\usepackage{microtype}      % microtypography
\usepackage{lipsum}
\usepackage{graphicx}

\usepackage{xcolor}
\usepackage{multirow}
\usepackage{array}
\usepackage{rotating} 
\usepackage{amssymb}
\usepackage{amsmath}
\usepackage[switch]{lineno}
\graphicspath{ {./images/} }

\title{Hierarchical Graph Capsule Network}

\author{Jinyu Yang\textsuperscript{\rm 1}\footnote{This work is done when Jinyu Yang works as an intern at Tencent AI Lab}, Peilin Zhao\textsuperscript{\rm 2}, Yu Rong\textsuperscript{\rm 2}, Chaochao Yan\textsuperscript{\rm 1}, Chunyuan Li\textsuperscript{\rm 1}, Hehuan Ma\textsuperscript{\rm 1}, \\ Junzhou Huang\textsuperscript{\rm 1} \\ % All authors must be in the same font size and format. Use \Large and \textbf to achieve this result when breaking a line
}

\affiliations{
	\textsuperscript{\rm 1}University of Texas at Arlington\\ 
	\textsuperscript{\rm 2}Tencent AI Lab \\
	jzhuang@uta.edu
}

\begin{document}
	%\linenumbers
	\maketitle
	
	\begin{abstract}
		Graph Neural Networks (GNNs) draw their strength from explicitly modeling the topological information of structured data. However, existing GNNs suffer from limited capability in capturing the hierarchical graph representation which plays an important role in graph classification. In this paper, we innovatively propose hierarchical graph capsule network (HGCN) that can jointly learn node embeddings and extract graph hierarchies. Specifically, disentangled graph capsules are established by identifying heterogeneous factors underlying each node, such that their instantiation parameters represent different properties of the same entity. To learn the hierarchical representation, HGCN characterizes the part-whole relationship between lower-level capsules (part) and higher-level capsules (whole) by explicitly considering the structure information among the parts. Experimental studies demonstrate the effectiveness of HGCN and the contribution of each component. Code: https://github.com/uta-smile/HGCN
		
	\end{abstract}
	
	%To learn the hierarchical representation, HGCN characterize the part-whole relationships and dynamically identify higher-level capsules (whole) that receive a cluster of similar predictions from lower-level capsules (part).
	\section{Introduction}
	GNNs \cite{scarselli2008graph}, especially graph convolutional networks \cite{bruna2013spectral,henaff2015deep} have demonstrated remarkable performance in modeling structured data in a wide variety of fields, such as social networks \cite{kipf2016semi,hamilton2017inductive, li2019semi} and graph-based representations of molecules \cite{gilmer2017neural,rong2020self}.
	The common practice is to recursively update node embeddings by aggregating (or message passing) information from topological neighbors such that the GNNs can capture the local structure of nodes. 
	Subsequently, the learned embeddings can be used in downstream analyses, e.g., node classification \cite{kipf2016semi,hamilton2017inductive,huang2018adaptive,xu2018representation,rong2019dropedge}, link prediction \cite{zhang2018link}, and graph classification \cite{duvenaud2015convolutional,dai2016discriminative,gilmer2017neural}.
	%However, those GNNs fail to capture hierarchical graph representations \cite{ying2018hierarchical}, which is especially problematic for graph classification where a graph-level representation is required.
	%For instance, in order to predict the properties of a given molecule, it would be highly desirable to encode the local structure of each atom by message passing \cite{gilmer2017neural}, and more importantly, to infer functional groups (a set of atoms) in a hierarchical manner.
	However, those GNNs fail to capture the hierarchical representations of graphs \cite{ying2018hierarchical}, which is essential for many scenarios. For instance, in order to predict the properties of a given molecule, it  would be highly desirable to infer the sub-parts which are important for the molecular properties hierarchically.
	
	To this end, various graph pooling methods are recently proposed, aiming to learn the coarse-grained graph structure by either reserving the most informative nodes \cite{gao2019graph,lee2019self} or aggregating nodes belonging to the same cluster \cite{ying2018hierarchical,yuanstructpool,khasahmadi2020memory,wang2019haarpooling,Bianchi2020spectral}. 
	In particular, the latter attracts considerable attention mainly attributed to its remarkable performance.
	Such kind of methods learn a cluster assignment matrix to map each node to a set of clusters that may correspond to strongly connected communities within a social network or functional groups within a molecule.
	%The intuition behind lies in that by iteratively coarsening the input graph, the complex hierarchical structure can be effectively learned and thereby leads to a more accurate graph representation.
	However, their limitations lie in that (i) simply grouping node features fails to effectively model the part-whole relationship that is crucial in characterizing the hierarchical structure, and (ii) they ignore the entanglement of the latent factors behind node embeddings, resulting in limited capacity in preserving detailed node/graph properties and modeling graph hierarchy. 
	For example, it is of particular importance to consider the interaction of heterogeneous factors (e.g., work, hobby) underlying each node, in order to identify the communities in a social network.
	
	Capsule neural networks (CapsNets) have proved its effectiveness in modeling hierarchical relationships on image data by exploiting the fact that while viewpoint changes have complicated effects on pixel intensities, they have linear effects at the part/object level \cite{sabour2017dynamic,hinton2018matrix,kosiorek2019stacked}.
	In contrast to convolutional neural networks, CapsNets use activity vectors or pose matrices to represent the entities. 
	Moreover, the viewpoint-invariant relationship between the part and the whole is characterized by trainable transformation matrices, which is under the assumption that the human visual system relies on parse tree-like structure to recognize objects.
	Such representations make CapsNets especially appealing in reasoning the part-whole hierarchy and robust to adversarial attacks \cite{hinton2018matrix,qin2019detecting}.
	However, how to effectively take advantage of CapsNets to benefit graph classification remains largely unexplored.
	
	In this work, we present the hierarchical graph capsule network (HGCN) that is able to jointly learn node embeddings and extract the hierarchical structure of the input graph.
	Specifically, to preserve detailed node/graph information, we build graph capsules by disentangling heterogeneous factors behind the node embeddings such that each capsule encodes different properties of the same entity.
	In order to capture the graph hierarchy, multiple graph capsule layers are stacked to get coarser and coarser representations.
	In each layer, (i) to infer the votes of instantiation parameters of higher-level capsules (wholes), we propose transformation GNNs to reason about the part-whole relationship by explicitly considering the structure information among lower-level capsules (parts);
	(ii) each of these votes that are weighted by a routing weight, are iteratively routed to the potential wholes that correspond to tight clusters in the votes.
	We further introduce the auxiliary graph reconstruction to enhance the representation capacity of graph capsules and the training stability.
	As a consequence, HGCN is capable of modeling hierarchical representations of the input graph and benefits the goal of graph classification. 
	
	Our main contributions can be summarized as:
	(i) we propose a novel capsule-based graph neural network to learn node embeddings and hierarchical graph representations simultaneously,
	(ii) we demonstrate the effectiveness of considering the entanglement of latent factors and the structure information within the parts in modeling part-whole relationships on the graph data;
	(iii) comprehensive empirical studies demonstrate that our method achieves remarkably superior improvement over the state-of-the-art approaches on 11 commonly used benchmarks.
	
	%contribution: not only what we did, but also highlight our advantages
	%None of them, however, XXX
	
	%Higher-level capsules ignore outliers, concentrating on clusters.
	%The outputs from one capsule (child) are routed to capsules in the next layer (parent) according to the child's ability to predict the parents' outputs. Over the course of a few iterations, each parents' outputs may converge with the predictions of some children and diverge from those of others, meaning that that parent is present or absent from the scene
	
	%For each possible parent, each child computes a prediction vector by multiplying its output by a weight matrix (trained by backpropagation).[3] Next the output of the parent is computed as the scalar product of a prediction with a coefficient representing the probability that this child belongs to that parent. A child whose predictions are relatively close to the resulting output successively increases the coefficient between that parent and child and decreases it for parents that it matches less well. After a few iterations, the coefficients strongly connect a parent to its most likely children, indicating that the presence of the children implies the presence of the parent in the scene
	
	%The coupling coefficients from a capsule i in layer l to all capsules in layer {l+1} sum to one
	
	\section{Related Work}
	
	\iffalse
	EigenPool introduced a graph pooling that used the local graph
	Fourier transform to extract subgraph information. Its potential drawback lies in the inherent computing bottleneck for the Laplacian-based graph Fourier transform, given the high
	computational cost for the eigendecomposition of the graph Laplacian.
	
	EigenPooling \cite{ma2019graph} summarize the subgraph information by also considering the subgraph structure; e local graph Fourier transform
	
	MinCutPool \cite{Bianchi2020spectral} spectral clustering; is a well-known clustering technique that leverages the Laplacian spectrum to find strongly connected communities on a graph; eigendecomposition of the Laplacian is expensive;  does not require to compute the spectral decomposition;
	
	Haar graph pooling \cite{wang2019haarpooling} (ICML 2020): compressive Haar transforms; it is computed by following a sequence of clusterings of the input graph; the compressive Haar transform filters out fine detail information in the Haar wavelet domain. 
	
	GMN \cite{khasahmadi2020memory}:  these models are not efficient as they require an iterative
	process of message passing after each pooling layer; cluster-based; assignment matrix-based
	\fi
	
	\subsubsection{Graph Neural Networks}
	GNNs attempt to exploit the structure information underlying graph structured data in order to benefit various downstream tasks \cite{li2015gated,kipf2016semi,hamilton2017inductive,li2018adaptive,xu2018representation,luan2019break,rong2019dropedge,you2019position}.  
	Recent studies have proved GNNs' wide applicability in, for example, drug discovery \cite{yan2020retroxpert,gilmer2017neural,ma2020multi}, protein interface prediction \cite{fout2017protein}, and recommendation system \cite{ying2018graph}.
	Current GNNs can be mainly summarized as two streams: spectral-based methods and spatial-based methods.
	The spectral-based methods largely rely on the convolution operation defined in the Fourier domain with spectral filters \cite{bruna2013spectral,henaff2015deep}.
	This kind of method is further simplified and extended by introducing polynomial spectral filters \cite{defferrard2016convolutional} or linear filters \cite{kipf2016semi}. 
	To deal with arbitrarily structured graphs, the spatial-based methods define convolutions directly on the graph by aggregating features from topological neighbors \cite{atwood2016diffusion,niepert2016learning,hamilton2017inductive,vaswani2017attention}. 
	%Specifically, \cite{velivckovic2017graph} incorporates the attention mechanism \cite{vaswani2017attention} into the propagation step to specify different weights to the neighbors. 
	
	\subsubsection{Graph Pooling}
	It is widely recognized that pooling operation plays an important role in graph classification \cite{errica2019fair} which requires the graph level representation.
	The most straightforward way is, to sum up, or take an average of all node features \cite{hamilton2017inductive,xu2018powerful}. 
	%An alternative is to sort node embeddings according to their structural roles and generate a sorted graph representation with a fixed size \cite{zhang2018end}.
	The limitation of such strategy is that the hierarchical information which is crucial in capturing graph structure is not considered.
	Inspired by the downsampling in convolutional neural networks, recent studies propose to adaptively keep the most informative nodes in a hierarchical manner \cite{gao2019graph,lee2019self}, or aggregate maximal cliques by only using topological information \cite{luzhnica2019clique}.
	%These methods, however, suffer from information loss and give rise to impaired graph representations.
	Another line of work focuses on finding strongly connected communities on a graph. 
	This is typically achieved by learning a cluster assignment matrix in order to map each node to a set of clusters \cite{ying2018hierarchical,ranjan2019asap,yuanstructpool,khasahmadi2020memory}. 
	Most recent studies approach this problem by leveraging more advanced clustering techniques, such as local graph Fourier transform \cite{ma2019graph}, spectral clustering \cite{Bianchi2020spectral}, and compressive Haar transforms \cite{wang2019haarpooling}.
	However, simply grouping node features has limited capacity in modeling the part-whole relationships, especially for biological data.
	In this work, we reason about the part-whole hierarchy by exploring the interaction of underlying latent factors and structure information among the parts, then use an routing mechanism to assign the parts to wholes.
	%each lower-level capsule (part) first makes predictions for the instantiation parameters of each higher-level capsule (whole). 
	%These predictions are then iteratively grouped using an routing mechanism to assign the parts to wholes.
	% guided by top-down feedback
	%existing cluster-based pooling methods represent a special case of our method when the transformation matrix is just an identity matrix.
	
	\subsubsection{Capsule Networks}
	A capsule \cite{hinton2011transforming} is a group of neurons whose orientation represents the instantiation parameters such as pose (position, size) of an entity (e.g., an object). 
	The probability that the entity exists can be represented by the capsule length \cite{sabour2017dynamic} or a logistic unit \cite{hinton2018matrix,kosiorek2019stacked}.
	Compared to a single neuron, a capsule contains different properties of the same entity and can preserve hierarchical relationships between lower-level capsules (e.g., eyes, mouth) and higher-level capsules (e.g., face). 
	Such part-whole relationships are described by trainable transformation matrices which are viewpoint-invariant. 
	Concretely, a lower-level capsule (part) makes predictions for the pose of each higher-level capsule (whole) by multiplying its own pose by the transformation matrices.
	Routing-by-agreement is then performed between two adjacent capsule layers to update the probability with which a part is assigned to a whole \cite{sabour2017dynamic,hinton2018matrix}.
	
	Inspired by this, GCAPS-CNN \cite{verma2018graph} builds capsules on graphs by considering higher-order statistical moments as instantiation parameters.
	Normal graph convolution is then carried out to aggregate information from neighbors and the covariance is computed as the permutation invariant feature for graph classification.
	Its most obvious drawback lies in that the hierarchical structure of the graph is not considered. 
	Different from GCAPS-CNN, we explicitly take into account the hierarchy between two consecutive capsule layers through trainable transformation GNNs.
	CapsGNN \cite{xinyi2018capsule} uses multiple GNN channels to build graph capsules and follow the same voting strategy as \cite{sabour2017dynamic} to predict higher-level capsules. 
	However, simply using the transformation matrix ignores the local structure information among lower-level capsules and fails to describe part-whole relationships in graphs. 
	Our method introduces transformation GNNs to reason about the pose of each whole in the layer above, which is in contrast to each individual part making its own prediction.
	A further advantage of transformation GNNs is that they save orders of magnitude of memory compared to transformation matrices used in previous work \cite{xinyi2018capsule,sabour2017dynamic,hinton2018matrix}.
	Furthermore, different from CapsGNN that reconstructs the histogram of input nodes, we reconstruct the adjacency matrix of the input graph to ensure the quality of graph capsules and enhance the training stability.
	% 
	%As opposed to simply aggregate node features from neighbors, in this paper, we introduce graph capsule 
	
	\iffalse
	\paragraph{Disentangled Representation Learning}
	We dynamically disentangle the node representations
	1. PAYLESS ATTENTION WITH LIGHTWEIGHT AND DYNAMIC CONVOLUTIONS 
	2. dynamic convolution: attention over convolution kernels
	3. Disentangled graph convolutional networks \cite{ma2019disentangled}
	\fi
	
	\section{Preliminaries}
	\subsubsection{Graph Classification}
	A graph $G$ with $N$ nodes is represented as $(\mathbf{A},\mathbf{X})$, where $\mathbf{A}\in\{0,1\}^{N \times N}$ is the adjacency matrix, and $\mathbf{X} \in \mathbb{R}^{N \times d}$ is the node feature matrix with feature dimension $d$.
	Given a set of labeled graphs $\mathcal{D}=\{(G_1, y_1),(G_2, y_2),...\}$, the goal of graph classification is to learn a mapping $f:\mathcal{G} \rightarrow \mathcal{Y}$, where $G_i \in \mathcal{G}$ and $y_i \in \mathcal{Y}$.
	For example, each graph is a molecule, and its label indicates whether it is toxic.
	
	\subsubsection{Graph Neural Networks}
	To extract useful information from local neighborhoods, our method is built upon GNNs by following the general "message-passing" paradigm, which is formulated as:
	\begin{equation}
	\mathbf{H}^{(l+1)}=\mathcal{M}(\mathbf{A},\mathbf{H}^{(l)};\mathbf{W}^{(l)}),
	\end{equation}
	where $\mathcal{M}$ indicates the message passing function with various possible implementations \cite{kipf2016semi, hamilton2017inductive}, $\mathbf{W}^{(l)}$ is learnable weight matrix, $\mathbf{H}^{(l+1)}$ and $\mathbf{H}^{(l)}$ are the node embeddings of layer $l+1$ and $l$, respectively.
	The input node embeddings $\mathbf{H}^{(1)}$ are initialized using the node feature $\mathbf{X}$, i.e., $\mathbf{H}^{(1)}=\mathbf{X}$.
	The final node representations are denoted by GNN$(\mathbf{A}, \mathbf{X})=\mathbf{H}^{(L_\text{GNN})}\in \mathbb{R}^{N \times h}$ with $L_\text{GNN}$ iterations.

	\begin{figure*}[t]
		\begin{center}
			\includegraphics[width=1\linewidth]{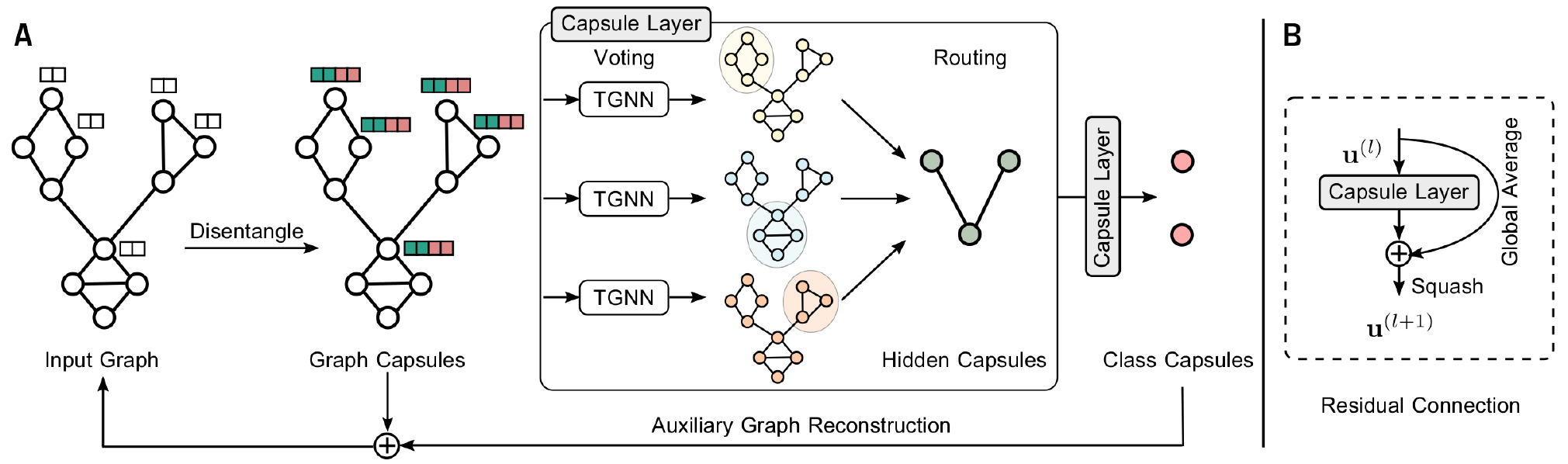}
		\end{center}
		\caption{An overview of the proposed framework. (A) Given an input graph, we build graph capsules by learning disentangled node representations in order to take into account the heterogeneous factors behind each node. TGNNs are established to characterize the part-whole relationship, and a routing strategy is used to predict higher-level capsules that receive a cluster of similar votes. (B) The residual connection that combines fine, low layer information with coarse, high layer information. }
		\label{fig:framework}
		%\vspace{-0.2in}
	\end{figure*}
	%The last capsule layer is designed for reasoning about the class capsules whose number depends on the class labels
	
	\begin{figure}
		\begin{center}
			\includegraphics[width=0.85\linewidth]{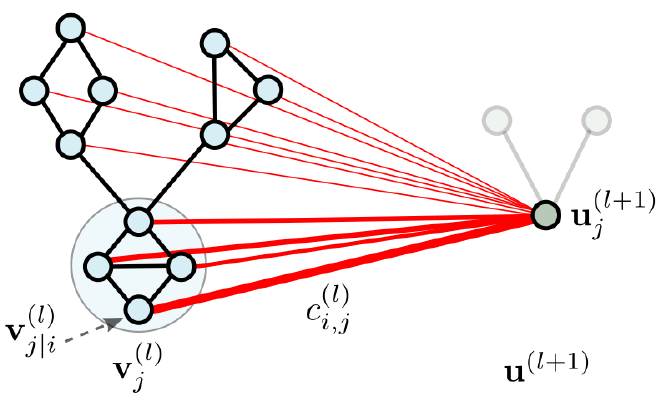}
		\end{center}
		\caption{Cluster by agreement.}
		\label{fig:routing}
		%\vspace{-0.2in}
	\end{figure}

	\section{Methodology}
	In this section, we begin by first briefing the proposed HGCN as shown in Figure~\ref{fig:framework}, then we detail each component in the following sections.
	The goal of HGCN is to jointly learn node embeddings and coarsen the graph through exploiting hierarchical information.
	To achieve this, we disentangle node representations to build the graph capsule by considering heterogeneous factors underlying each edge connection.
	Therefore, each graph capsule is composed of multiple independent latent factors that represent different properties of the same entity.
	To learn hierarchical graph representations, transformation GNNs (TGNNs) are proposed to encode the part-whole relationship between lower-level and higher-level graph capsules.
	Specifically, a capsule in one layer votes for the instantiation parameters of each capsule in the layer above through TGNNs which highly depend on the structure information of lower-level capsules.
	Each of these votes is then routed to a higher-level capsule that receives a cluster of similar votes by a routing-by-agreement strategy.
	To encourage the graph capsules to encode the instantiation parameters of the input graph and also enhance the training stability, we further introduce the auxiliary graph reconstruction to reconstruct the input adjacency matrix.

	\subsection{Disentangled Graph Capsules}
	In most cases, highly complex interactions are involved in the connection between each node pair in a graph. 
	For example, the edges between a node and its neighbors in a social network are driven by heterogeneous factors, since a person connects with others for various reasons such as exercise, work, etc. 
	Therefore, it is necessary to disentangle the explanatory factors of variations underlying the node representations. 
	Furthermore, each node embedding is considered as multiple individual scalar features in existing GNNs, which are proved to have limited capability in preserving the graph properties \cite{verma2018graph,xinyi2018capsule}. 
	
	To address these two limitations, motivated by \cite{sabour2017dynamic}, we propose graph capsules to describe the given graph.
	Specifically, we disentangle the latent factors of each node embedding and use the disentangled node representation to represent graph capsules (Figure~\ref{fig:framework}A). 
	In this way, each graph capsule is composed of multiple heterogeneous factors, and each factor describes a specific instantiation parameter of the entity/node. 
	Formally, given $G=(\mathbf{A}, \mathbf{X})$, the node $i$ is denoted by $\mathbf{x}_i \in \mathbb{R}^d$. 
	We project the input node features into $K$ different subspaces, assuming that there are $K$ latent factors/instantiation parameters:
	\begin{equation}
	\mathbf{z}_{i,k}=\sigma (\mathbf{W}_k^T \mathbf{x}_i) + \mathbf{b}_k,
	\end{equation}
	where $\mathbf{W}_k \in \mathbb{R}^{d \times \frac{h}{K}}$ and $\mathbf{b}_k \in \mathbb{R}^{\frac{h}{K}}$ are learnable parameters, $\sigma$ is a nonlinear activation function, and $\frac{h}{K}$ is the dimension of each factor.  
	Although more sophisticated implementations of node disentanglement are possible \cite{ma2019disentangled}, we use linear projection in our study attributed to its efficiency and remarkable performance.
	Therefore, each graph capsule is represented by a pose matrix $\mathbf{Z}_i \in \mathbb{R}^{K \times \frac{h}{K}}$ \cite{hinton2018matrix}. For simplicity, we reshape $\mathbf{Z}_i$ to the vector format $\mathbf{z}_i \in \mathbb{R}^h$.
	Recall that the existence probability of an entity represented by a capsule is measured by the capsule length \cite{sabour2017dynamic}, we thus squash $\mathbf{z_i}$ as follows:
	\begin{equation}
	{\mathbf{p}}_{i}= squash(\mathbf{z}_{i}) = \frac{\|\mathbf{z}_{i}\|^2}{1+\|\mathbf{z}_{i}\|^2} \frac{\mathbf{z}_{i}}{\|\mathbf{z}_{i}\|},
	\end{equation}
	where $\mathbf{u}_i^{(1)} = {\mathbf{p}}_{i} \in \mathbb{R}^{h}$ is the primary graph capsule representing the lowest level of entities, such as atoms in the molecular graph. 
	%It is noteworthy that ${\mathbf{P}}_{i}$ describes $K$ latent factors underlying node $i$ and is more informative in preserving the node/graph properties than $\mathbf{X}_{i}$.
	
	%Each primary capsule is squashed by a squash function. The length of a capsule indicates the probability that the entity exists.
	%However, the hierarchical relationship between heterogeneous factors of lower-level nodes and heterogeneous factors of higher-level nodes (or clusters) are not considered in existing graph pooling methods.
	%Try conditional or dynamic disentangled node representation, but not work well (ablation study). 
	
	\subsection{Hierarchical Capsule Layers}
	
	To obtain hierarchical graph representation, it is essential to capture the part-whole relationship between adjacent capsule layers.
	Such relationship is measured by viewpoint-invariant transformation matrix $\mathbf{T}^{(l)}_{i,j} \in \mathbb{R}^{d_{l} \times d_{l+1}}$ for each pair of lower-level capsule $\mathbf{u}^{(l)}_i$ and higher-level capsule $\mathbf{u}^{(l+1)}_j$ in previous studies \cite{sabour2017dynamic,hinton2018matrix}, where $d_l$ and $d_{l+1}$ are the capsule dimensions of $\mathbf{u}^{(l)}_i$ and $\mathbf{u}^{(l+1)}_j$, respectively. 
	However, $\mathbf{T}^{(l)}_{i,j}$ totally ignores the structure information within $\mathbf{u}^{(l)}$, which is especially problematic for graph structured data.
	Furthermore, $\mathbf{T}^{(l)} \in \mathbb{R}^{N_{l} \times N_{l+1} \times d_{l} \times d_{l+1}}$ is extremely memory-consuming for the scenario where a large number of high-dimensional capsules are required.
	
	To overcome these difficulties, we propose the transformation GNNs (TGNNs) to vote for the instantiation parameters of higher-level graph capsules (\textbf{voting}). 
	When multiple votes agree, a higher-level capsule that receives a cluster of similar pose votes becomes active (\textbf{routing}).
	More concretely, we denote the graph capsules at layer $l$ as $\mathbf{u}^{(l)} \in \mathbb{R}^{N_l \times d_l}$, the capsule number as $N_l$, and the adjacency matrix as $\mathbf{A}^{(l)}$.
	Our goal is to decide which capsules to activate in $\mathbf{u}^{(l+1)} \in \mathbb{R}^{N_{l+1} \times d_{l+1}}$ and how to assign each active lower-level capsule $\mathbf{u}^{(l)}_i$ to one active higher-level capsule $\mathbf{u}^{(l+1)}_j$.
	In practice, we set $N_{l+1} < N_l$ in order to get coarser and coarser graph representations (Figure~\ref{fig:framework}A).
	%We initialize $\mathbf{u}_i^{(1)}$ as $\mathbf{P}_i$.
	
	\subsubsection{Voting} For all capsules in $\mathbf{u}^{(l)}$, their poses are transformed by TGNNs to cast votes for the pose of each capsule in $\mathbf{u}^{(l+1)}$ by the following equation,
	\begin{equation}
	\mathbf{v}^{(l)}_j = \text{TGNN}_j(\mathbf{A}^{(l)}, \mathbf{u}^{(l)}),
	\end{equation}
	where ${\mathbf{v}}^{(l)}_j \in \mathbb{R}^{N_l \times d_{l+1}}$. Specifically, ${\mathbf{v}}^{(l)}_{j|i} \in \mathbb{R}^{d_{l+1}}$ is the vote for the pose of $\mathbf{u}^{(l+1)}_j$ predicted by the capsule $\mathbf{u}^{(l)}_i$.
	Note, TGNNs are learned discriminatively and could learn to represent part-whole relationships by considering the structure information of capsules in $\mathbf{u}^{(l)}$.
	This is different from previous studies that use one transformation matrix for each pair of $(\mathbf{u}^{(l)}_i, \mathbf{u}^{(l+1)}_j)$. 
	Compared to transformation matrices, TGNNs also save $N_l$ orders of magnitude memory.
	
	\subsubsection{Routing} Each of these votes is then weighted by an routing weight $c^{(l)}_{i,j}$ with which a part is assigned to a whole, where $c^{(l)}_{i,j} \geqslant 0$ and $\sum_{j=1}^{N_{l+1}} c^{(l)}_{i,j}=1$.
	Here, $c^{(l)}_{i,j}$ is iteratively updated using an "routing-by-agreement" mechanism such that each vote in ${\mathbf{v}}^{(l)}$ is routed to a capsule in $\mathbf{u}^{(l+1)}$ that receives a cluster of similar votes (Figure~\ref{fig:routing}).
	Formally, $c^{(l)}_{i,j}$ is defined as $c^{(l)}_{i,j} = {\text{exp}(b^{(l)}_{i,j})}/{\sum_k \text{exp}(b^{(l)}_{i,k})}$, where $b^{(l)}_{i,j}$ is initialized as $b^{(l)}_{i,j}=0$.
	To iteratively search for the vote cluster, in each iteration we have, 
	\begin{equation}
	\mathbf{u}^{(l+1)}_j = squash(\sum_i c^{(l)}_{i,j}\mathbf{v}^{(l)}_{j|i})
	\end{equation}
	where $\mathbf{u}^{(l+1)}_j$ is the predicted capsule $j$ in layer $l+1$, representing a tight cluster of votes from layer $l$. 
	We update $b^{(l)}_{i,j}$ with $b^{(l)}_{i,j} = b^{(l)}_{i,j} + a^{(l)}_{i,j}$, where $a^{(l)}_{i,j} = \mathbf{v}^{(l)}_{j|i} \cdot \mathbf{u}^{(l+1)}_j$ indicates the agreement between each vote and vote cluster. 
	It is worth mentioning that such top-down feedback also has a beneficial effect on the aggregation in the proposed TGNNs, such that TGNNs can more focus on aggregating information from neighbors that are likely to be in the same cluster.
	After $R$ iterations, we get higher-level graph capsules $\mathbf{u}^{(l+1)}$ and the coarsened adjacency matrix defined as $\mathbf{A}^{(l+1)}={\mathbf{C}^{(l)}}^T \mathbf{A}^{(l)} \mathbf{C}^{(l)} \in \mathbb{R}^{N_{l+1} \times N_{l+1}}$.
	As opposed to generating structurally independent higher-level capsules in previous work \cite{sabour2017dynamic, xinyi2018capsule}, the capsule layer we developed is able to explicitly preserve the structure information which is encoded in $\mathbf{A}^{(l+1)}$.
	
	Drawing inspiration from \cite{long2015fully}, we add a residual connection at each pair of consecutive capsule layers, aiming to provide fine-grained information to higher-level capsules (Figure~\ref{fig:framework}B).
	Formally, we have $\mathbf{u}^{(l+1)} \leftarrow \mathbf{u}^{(l+1)} + \text{GA}(\mathbf{u}^{(l)})$, where $\text{GA}$ indicates the global average operation. 
	
	By stacking multiple capsule layers, we get the class capsules $\mathbf{u}^{(L)} \in \mathbb{R}^{O \times d_L}$ which are intended to encode feature attributes corresponding to the class, where $O$ is the number of graph categories.
	The classification loss is measured by a margin loss \cite{sabour2017dynamic} which is formulated as:
	\begin{equation}
	\begin{aligned}
	\mathcal{L}_{m}(\mathbf{A}, \mathbf{X}) = {} &
	\sum\limits_{o \in O} [T_o \hspace{0.1cm} \text{max}(0, m^+ - \|\mathbf{u}^L_o\|)^2 + \\ & \lambda(1-T_o) \hspace{0.1cm} \text{max}(0, \|\mathbf{u}^L_o\|-m^-)^2],
	\end{aligned}
	\end{equation}
	where $m^+=0.9$, $m^-=0.1$, $T_o=1$ iff the input graph has label $o$, and $\lambda$ is ued to stop the initial learning from shrinking the length of class capsules.

	\subsection{Auxiliary Graph Reconstruction} 
	To encourage the graph capsules to encode the instantiation parameters of the input graph and to improve the training stability, we introduce a reconstruction loss to constraint the capsule reconstruction to closely match the class-conditional distribution.
	Specifically, we first mask out all but the winning capsule (the capsule corresponds to ground truth) and combine them with primary capsules by following the equation,
	\begin{equation}
	\mathbf{Z} = \mathbf{u}^{(1)} + (\mathbf{W}_r^T \Phi(\mathbf{u}^{(L)}) + \mathbf{b}_r),
	\end{equation}
	where $\mathbf{Z} \in \mathbb{R}^{N \times d_1}$, $\Phi$ is the mask operation, $\mathbf{W}_r \in \mathbb{R}^{(O \times d_L) \times d_1}$, and $\mathbf{b}_r \in \mathbb{R}^{d_1}$. 
	The reconstruction loss is then defined as,
	\begin{equation}
	\mathcal{L}_{r}(\mathbf{A}, \mathbf{X}) = -\frac{1}{N^2}\sum_{j=1}^N \sum_{k=1}^N\sum_{c\in\{0,1\}} \mathbf{A}^{(j,k,c)}log(\mathbf{\tilde{A}}^{(j,k,c)}),
	\end{equation}
	where $\mathbf{\tilde{A}}=\mathbf{Z} \mathbf{Z}^T$ is the reconstructed adjacency matrix of the input graph.
	Taken together, we reach the optimization objective of our method as $\min\limits_{\theta} \sum\limits_{G \in \mathcal{D}} \mathcal{L}_m(\mathbf{A}, \mathbf{X}) + \beta \mathcal{L}_r(\mathbf{A}, \mathbf{X})$, where $\theta$ are all learnable parameters, and $\beta$ leverages the importance of $\mathcal{L}_r$.
	The whole training process is detailed in Algorithm~\ref{alg:alg_1}.
	\setlength{\textfloatsep}{10pt}%
	\begin{algorithm}[t]
		\SetAlgoLined
		\textbf{Input:} $G=(\mathbf{A}, \mathbf{X}), \mathbf{A} \in \mathbb{R}^{N \times N}, \mathbf{X} \in \mathbb{R}^{N \times d}$ \\
		\KwResult{class capsules $\mathbf{u}^{(L)}$}
		\For{$i\leftarrow 1$ \KwTo $N$}{
			\For{$k\leftarrow 1$ \KwTo $K$}{
				$\mathbf{z}_{i,k}=\sigma (\mathbf{W}_k^T \mathbf{x}_i) + \mathbf{b}_k$ \hspace{13pt} \/// $K$ latent factors
			}
		}
		$\mathbf{u}^{(1)}_i=squash(\mathbf{z}_i)$ \hspace{14pt} \/// disentangled graph capsules
		
		\For{$l\leftarrow 1$ \KwTo $L$}{
			$b^{(l)}_{i,j} = 0$ \\
			\For{$j\leftarrow 1$ \KwTo $N_{l+1}$}{
				$\mathbf{v}^{(l)}_j = \text{TGNN}_j(\mathbf{A}^{(l)}, \mathbf{u}^{(l)})$ \hspace{1.5cm} \/// votes \\
			}
			\For{$r\leftarrow 1$ \KwTo $R$}{
				$c^{(l)}_{i,j} = {\text{exp}(b^{(l)}_{i,j})}/{\sum_k \text{exp}(b^{(l)}_{i,k})}$  \\
				$\mathbf{u}^{(l+1)}_j = squash(\sum_i c^{(l)}_{i,j}\mathbf{v}^{(l)}_{j|i})$ \\
				$b^{(l)}_{i,j} = b^{(l)}_{i,j} + \mathbf{v}^{(l)}_{j|i} \cdot \mathbf{u}^{(l+1)}_j$
			}
			$\mathbf{A}^{(l+1)}={\mathbf{C}^{(l)}}^T \mathbf{A}^{(l)} \mathbf{C}^{(l)} \in \mathbb{R}^{N_{l+1} \times N_{l+1}}$
		}
		\textbf{return} $\mathbf{u}^{(L)} \in \mathbb{R}^{O \times d_L}$ \hspace{2cm} \/// class capsules
		
		\caption{Training process with $K$ latent factors, $L$ capsule layers, and $R$ iterations of routing.}
		\label{alg:alg_1}
	\end{algorithm}

	\begin{table*}
		\footnotesize
		\setlength\tabcolsep{4.5pt}
		\begin{center}
			\begin{tabular}{ @{} l|c|*{7}{c} @{} }
				\toprule
				%\multicolumn{24}{ c }{\bf GTA5$\,\to\,$Cityscapes } \\
				%\midrule
				\bf Algorithm & & \bf MUTAG & \bf NCI1 & \bf PROTEINS & \bf D\&D & \bf ENZYMES & \bf PTC & \bf NCI109 \\
				\midrule
				
				AWE & \multirow{4}{*}{\rotatebox[origin=c]{90}{Kernel}} &
				87.87$\pm$9.76 & \textemdash & \textemdash & 71.51$\pm$4.02 & 35.77$\pm$5.93 & \textemdash &\textemdash \\
				
				GK & & 
				81.58$\pm$2.11 & 62.49$\pm$0.27 & 71.67$\pm$0.55 & 78.45$\pm$0.26 & 32.70$\pm$1.20 & 59.65$\pm$0.31 & 62.60$\pm$0.19 \\
				
				WL & & 
				82.05$\pm$0.36 & 82.19$\pm$0.18 & 74.68$\pm$0.49 & 79.78$\pm$0.36 & 52.22$\pm$1.26 & 57.97$\pm$0.49 & 82.46$\pm$0.24 \\
				
				DGK & & 
				87.44$\pm$2.72 & 80.31$\pm$0.46 & 75.68$\pm$0.54 & 73.50$\pm$1.01 & 53.43$\pm$0.91 & 60.08$\pm$2.55 & 80.32$\pm$0.33 \\

				\midrule
				SAGPool & 
				\multirow{11}{*}{\rotatebox[origin=c]{90}{GNN}} &
				\textemdash & 67.45$\pm$1.11 & 71.86$\pm$0.97 & 76.45$\pm$0.97 & \textemdash & \textemdash & 67.86$\pm$1.41 \\
				
				CLIQUEPOOL & &
				\textemdash & \textemdash & 72.59 & 77.33 & 60.71 & \textemdash & \textemdash \\
				
				ASAP & &
				\textemdash & 71.48$\pm$0.42 &  74.19$\pm$0.79 & 76.87$\pm$0.7 & \textemdash &
				\textemdash & 70.07$\pm$0.55 \\
				
				%HaarPool & & 
				%90.00$\pm$3.60 & 78.60$\pm$0.50 & 80.40$\pm$1.80 & \textemdash & \textemdash & \textemdash & 75.60$\pm$1.20 \\
				
				HaarPool & & 
				77.60$\pm$8.94 & 80.17$\pm$2.29 & 73.23$\pm$2.51 & \textemdash & \textemdash & \textemdash & 69.61$\pm$1.49 \\
				
				EigenPooling & & 
				79.50 & 77.00 & 76.60 & 78.60 & 64.50 & \textemdash & 74.90 \\
				
				DGCNN & & 
				85.83$\pm$1.66 & 74.44$\pm$0.47 & 75.54$\pm$0.94 & 79.37$\pm$0.94 & 51.00$\pm$7.29 & 58.59$\pm$2.47 & 75.03$\pm$1.72 \\
				
				PSCN & &  
				88.95$\pm$4.37 & 76.34$\pm$1.68 & 75.00$\pm$2.51 & 76.27$\pm$2.64 & \textemdash &
				62.29$\pm$5.68 & \textemdash \\				
				
				GIN & & 
				89.40$\pm$5.60 & 82.70$\pm$1.70 & 76.20$\pm$2.80 & \textemdash & \textemdash &
				64.60$\pm$7.00 & \textemdash \\								
				
				DIFFPOOL & & 
				\textemdash & \textemdash & 76.25 & 80.64 & 62.53 & \textemdash & 74.10 \\
				
				GCN & &
				87.20$\pm$5.11 & 83.65$\pm$1.69 & 75.65$\pm$3.24 & 79.12$\pm$3.07 & 66.50$\pm$6.91 & \textemdash & 70.70 \\
				
				GFN & & 
				90.84$\pm$7.22 & 82.77$\pm$1.49 & 76.46$\pm$4.06 & 78.78$\pm$3.49 & 70.17$\pm$5.58 & \textemdash & \textemdash \\					
				
				\midrule
				CapsGNN & 
				\multirow{3}{*}{\rotatebox[origin=c]{90}{Caps}} & 
				86.67$\pm$6.88 & 78.35$\pm$1.55 & 76.28$\pm$3.63 & 75.38$\pm$4.17 & 54.67$\pm$5.67 & \textemdash & \textemdash \\
				
				GCAPS-CNN & & 
				\textemdash & 82.72$\pm$2.38 & 76.40$\pm$4.17 & 77.62$\pm$4.99 & 61.83$\pm$5.39 &
				66.01$\pm$5.91 & 81.12$\pm$1.28 \\
				
				\textbf{Ours} & & 
				\textbf{93.16$\pm$6.10} & \textbf{84.87$\pm$1.68} & \textbf{77.99$\pm$3.16} & \textbf{80.99$\pm$2.58} & \textbf{78.00$\pm$4.89} & \textbf{66.54$\pm$7.97} & \textbf{83.91$\pm$1.27} \\
				
				\bottomrule
			\end{tabular}
		\end{center}
		\caption{Performance comparison on biological graphs. "Caps" indicates capsule-based GNNs.}
		\label{table:biological_c1}
		%\vspace{-0.2in}
	\end{table*}
	
	\iffalse
	\begin{table}
		\caption{Performance comparison on biological graphs ($C^*$).}
		\label{table:biological_c2}
		
		\footnotesize
		\setlength\tabcolsep{4.5pt}
		\begin{center}
			\begin{tabular}{ @{} l|c|*{3}{c} @{} }
				\toprule
				%\multicolumn{24}{ c }{\bf GTA5$\,\to\,$Cityscapes } \\
				%\midrule
				\bf Algorithm & & \bf PROTEINS & \bf D\&D & \bf ENZYMES \\
				\midrule
				
				STRUCTPOOL &
				\multirow{4}{*}{\rotatebox[origin=c]{90}{GNN}} &
				80.36 & 84.19 & 63.83 \\
				
				MemGNN & & 
				81.35 & 82.92 & 75.50 \\
				
				GMN & & 
				\textbf{82.25} &  \textbf{84.40} & 78.66 \\
				
				\textbf{Ours} & &
				81.32$\pm$2.23 & 84.04$\pm$2.38 & \textbf{84.17$\pm$4.03} \\
				
				\bottomrule
			\end{tabular}
		\end{center}
		%\vspace{-0.2in}
	\end{table}
	\fi

	\begin{table}
		\footnotesize
		\setlength\tabcolsep{1.5pt}
		\begin{center}
			\begin{tabular}{ @{} l|c|*{4}{c} @{} }
				\toprule
				%\multicolumn{24}{ c }{\bf GTA5$\,\to\,$Cityscapes } \\
				%\midrule
				\bf Algorithm & & \bf COLLAB & \bf IMDB-B & \bf IMDB-M & \bf RE-B \\
				\midrule
				
				GK & \multirow{4}{*}{\rotatebox[origin=c]{90}{Kernel}} &
				72.84$\pm$0.28 & 65.87$\pm$0.98 & 43.89$\pm$0.38 & 65.87$\pm$0.98 \\
				
				AWE & &
				73.93$\pm$1.94 & 74.45$\pm$5.83 & 51.54$\pm$3.61 & 87.89$\pm$2.53 \\
				
				WL & & 
				79.02$\pm$1.77 & 73.40$\pm$4.63 & 49.33$\pm$4.75 & 81.10$\pm$1.90 \\
				
				DGK & &
				73.09$\pm$0.25 & 66.96$\pm$0.56 & 44.55$\pm$0.52 & 78.04$\pm$0.39 \\
				
				\midrule
				
				PSCN &  
				\multirow{6}{*}{\rotatebox[origin=c]{90}{GNN}} &
				72.60$\pm$2.15 & 71.00$\pm$2.29 & 45.23$\pm$2.84 & 86.30$\pm$1.58 \\
				
				DGCNN & &
				73.76$\pm$0.49 & 70.03$\pm$0.86 & 47.83$\pm$0.85 & 76.02$\pm$1.73 \\
				
				DIFFPOOL & & 
				75.48 & \textemdash & \textemdash & \textemdash \\						
				
				GCN & &
				81.72$\pm$1.64 & 73.30$\pm$5.29 & 51.20$\pm$5.13 & \textemdash \\
				
				GFN & & 
				81.50$\pm$2.42 & 73.00$\pm$4.35 & 51.80$\pm$5.16 & \textemdash \\
				
				GIN & & 
				80.20$\pm$1.90 & 75.10$\pm$5.10 & 52.30$\pm$2.80 & 92.40$\pm$2.50 \\
				
				\midrule
				GCAPS-CNN &  
				\multirow{3}{*}{\rotatebox[origin=c]{90}{Caps}} &
				77.71$\pm$2.51 & 71.69$\pm$3.40 & 48.50$\pm$4.10 & 87.61$\pm$2.51 \\
				
				CapsGNN & & 
				79.62$\pm$0.91 & 73.10$\pm$4.83 & 50.27$\pm$2.65 & \textemdash \\
				
				\textbf{Ours} & & 
				\textbf{82.86$\pm$1.81} & \textbf{77.20$\pm$4.73} & \textbf{52.80$\pm$2.45} & \textbf{93.15$\pm$1.58} \\
				
				\bottomrule
			\end{tabular}
		\end{center}
		\caption{Performance comparison on social graphs.}
		\label{table:social_c1}
		%\vspace{-0.2in}
	\end{table}
	
	\iffalse
	\begin{table}
		\caption{Performance comparison on social graphs ($C^*$).}
		\label{table:social_c2}
		
		\footnotesize
		\setlength\tabcolsep{1.4pt}
		\begin{center}
			\begin{tabular}{ @{} l|c|*{4}{c} @{} }
				\toprule
				%\multicolumn{24}{ c }{\bf GTA5$\,\to\,$Cityscapes } \\
				%\midrule
				\bf Algorithm & & \bf COLLAB & \bf IMDB-B & \bf IMDB-M & \bf RE-B \\
				\midrule
				
				STRUCTPOOL &
				\multirow{4}{*}{\rotatebox[origin=c]{90}{GNN}} & 
				74.22 & 74.70 & 52.47 & \textemdash \\
				
				MemGNN & &
				77.0 & \textemdash & \textemdash & 85.55  \\
				
				GMN & & 
				80.18 & \textemdash & \textemdash & 95.28 \\
				
				\textbf{Ours} & &
				\textbf{84.80$\pm$1.57} & \textbf{79.80$\pm$3.39} & \textbf{55.80$\pm$2.20} &
				\textbf{95.30$\pm$1.30} \\
				
				\bottomrule
			\end{tabular}
		\end{center}
		%\vspace{-0.2in}
	\end{table}
	\fi

	\section{Experiments}
	In this section, we conduct empirical studies on 11 benchmark datasets and demonstrate HGCN's superiority over a number of state-of-the-art graph classification methods.
	Extensive ablation studies are also performed to evaluate the effectiveness of each component in our model.
	
	\subsubsection{Datasets}
	Eleven commonly used benchmarks including (i) seven biological graph datasets, i.e., MUTAG, NCI1, PROTEINS, D\&D, ENZYMES, PTC, NCI109; and (ii) four social graph datasets, i.e., COLLAB, IMDB-Binary (IMDB-B), IMDB-Multi (IMDB-M), Reddit-BINARY (RE-B), are used in this study. 
	It is noteworthy that the social graphs have no node attributes, while the biological graphs come with categorical node attributes. 
	More details about the data statistics and properties can be found in Supplementary.
	
	\subsubsection{Baseline Methods}
	We compare with two capsule-based methods, i.e., CapsGNN \cite{xinyi2018capsule} and GCAPS-CNN \cite{verma2018graph}. 
	We also conduct a comparison with a number of state-of-the-art GNN-based methods, including PATCHY-SAN (PSCN) \cite{niepert2016learning}, PSCN \cite{niepert2016learning}, GCN \cite{kipf2016semi}, Deep Graph CNN (DGCNN) \cite{zhang2018end}, CLIQUEPOOL \cite{luzhnica2019clique}, DIFFPOOL \cite{ying2018hierarchical}, ASAP \cite{ranjan2019asap}, SAGPool \cite{lee2019self}, EigenPooling \cite{ma2019graph}, GIN \cite{xu2018powerful}, GFN \cite{chen2019powerful}, HaarPool \cite{wang2019haarpooling}, STRUCTPOOL \cite{yuanstructpool}, and MemGNN/GMN \cite{khasahmadi2020memory}.
	For kernel-based methods, we consider WL \cite{shervashidze2011weisfeiler}, DGK \cite{yanardag2015deep}, AWE \cite{ivanov2018anonymous}, and GK \cite{shervashidze2009efficient}.
	%MinCutPool \cite{Bianchi2020spectral}
	%if Val loss < best val loss:
	%  change the patience
	%save the current model
	%After training, use the saved model to test the performance on the test data. Then average the accuracy over ten folds. This is not a ten-fold cross validation, but just randomly split the data for ten times.
	
	%HaarPool \cite{wang2019haarpooling}split the data into train/validation/test (8/1/1); if the validation loss < minimum validation loss, save the current model; early stoppling and then test the saved mode on the test data. They repeat all experiments ten times with different random seeds. No cross validation.
	
	\subsubsection{Experimental Settings}
	We set $K=4$, $R=3$, $\lambda=0.5$, $\beta=0.1$, $L=2$, and
	follow the same settings in previous studies \cite{ying2018hierarchical} to perform 10-fold cross-validation for performance evaluation. 
	For each dataset, we select a single epoch that has the best cross-validation accuracy averaged over the 10 folds, and report the average and standard deviation of test accuracies at the selected epoch.
	HaarPool \cite{wang2019haarpooling} repeats each experiment 10 times with different random seeds, for a fair comparison, we run HaarPool with 10-fold cross-validation and report the result.
	STRUCTPOOL \cite{yuanstructpool} and MemGNN/GMN \cite{khasahmadi2020memory} use a different assessment criterion that selects the epoch with the best test accuracy on each fold and then take the average.
	We name such assessment criterion as $C^*$ and run our method by following the same paradigm for comparison. 
	Unless otherwise indicated, we use the result reported in the original paper for other baseline methods.
	For TGNNs, we adopt the GCN \cite{kipf2016semi} with $L_\text{GNN}=1$.
	%More details are referred to Supplementary.
	
	%Two assessment criteria are used in our study: (C1) select a single epoch that has the best cross-validation accuracy averaged over the 10 folds; and (C2) select the epoch with the best accuracy on each fold and then take the average.
	\iffalse
	\begin{figure}
		\begin{center}
			\includegraphics[width=0.9\linewidth]{Figure/visualize.eps}
		\end{center}
		\caption{Visualization of routing weights.}
		\label{fig:visualization}
		%\vspace{-0.1in}
	\end{figure}
	\fi
	
	\begin{table*}
		\footnotesize
		\setlength\tabcolsep{4.5pt}
		\begin{center}
			\begin{tabular}{ @{} l|c|*{7}{c} @{} }
				\toprule
				%\multicolumn{24}{ c }{\bf GTA5$\,\to\,$Cityscapes } \\
				%\midrule
				& & \bf MUTAG & \bf NCI1 & \bf PROTEINS & \bf D\&D & \bf ENZYMES & \bf PTC & \bf NCI109 \\
				\midrule
				
				A1 &  
				\multirow{3}{*}{\rotatebox[origin=c]{90}{ablation}} &
				91.46$\pm$5.77 & 80.24$\pm$1.78 & 76.37$\pm$3.11 & 78.35$\pm$2.73 & 70.00$\pm$4.51 & 63.61$\pm$12.47 & 80.64$\pm$2.09 \\
				
				A2 &
				& 92.08$\pm$5.10 & \textbf{85.28$\pm$1.37} & 77.63$\pm$3.03 & 80.64$\pm$3.65 & 77.00$\pm$5.82 & 64.24$\pm$8.45 & 83.84$\pm$1.41 \\
				
				A3 &
				& 92.11$\pm$7.13 & 84.87$\pm$1.07 & 77.54$\pm$3.44 & 79.96$\pm$3.26 & 77.67$\pm$3.70 & 65.10$\pm$8.81 & 83.84$\pm$1.48 \\
				
				%A3b &
				%& 92.63$\pm$5.66 & 84.94$\pm$1.05 & 77.63$\pm$2.42 & 81.06$\pm$2.88 & %\textbf{78.17$\pm$5.85} & 63.68$\pm$6.81 & 84.08$\pm$1.57\\
				\midrule
				\textbf{Ours} &
				& \textbf{93.16$\pm$6.10} & 84.87$\pm$1.68 & \textbf{77.99$\pm$3.16} & \textbf{80.99$\pm$2.58} & \textbf{78.00$\pm$4.89} & \textbf{66.54$\pm$7.97} & \textbf{83.91$\pm$1.27} \\
				\midrule
				
				%CondDisen &  
				%\multirow{2}{*}{\rotatebox[origin=c]{90}{$f_d$}} &
				%92.11$\pm$6.68 & 83.02$\pm$1.79 & 76.82$\pm$2.79 & 79.29$\pm$2.69 & %78.00$\pm$6.66 & 65.97$\pm$9.82 & 81.32$\pm$2.03 \\
				
				%DynaDisen &
				%& 92.63$\pm$6.18 & 83.77$\pm$1.23 & 77.27$\pm$2.25 & 79.80$\pm$2.53 & %\textbf{78.17$\pm$6.45} & 66.26$\pm$7.77 &
				%82.87$\pm$1.54 \\
				\midrule
				
				2 &  
				\multirow{2}{*}{\rotatebox[origin=c]{90}{$K$}} &
				91.55$\pm$5.64 & 84.50$\pm$1.87 & 77.90$\pm$3.31 & 80.04$\pm$2.77 & 77.50$\pm$5.05 & 64.81$\pm$9.45 & \textbf{84.15$\pm$1.93} \\
				
				8 &
				& 93.16$\pm$6.59 & 84.65$\pm$1.17 & 77.18$\pm$2.74 & 78.69$\pm$2.99 & 77.83$\pm$4.97 & 65.71$\pm$7.95 & 84.03$\pm$2.31 \\
				\midrule
				
				1 &  
				\multirow{4}{*}{\rotatebox[origin=c]{90}{$R$}} &			
				91.05$\pm$6.59 & 83.36$\pm$1.62 & 76.64$\pm$2.46 & 79.45$\pm$2.47 & 77.33$\pm$5.34 & 64.23$\pm$8.60 & 82.26$\pm$1.75 \\
				
				2 &
				& 92.11$\pm$6.68 & 84.01$\pm$1.68 & 77.45$\pm$2.92 & 79.97$\pm$3.60 & 77.83$\pm$3.77 & 65.13$\pm$8.07 & 83.35$\pm$1.36 \\
				
				4 &
				& 92.63$\pm$5.66 & 84.26$\pm$1.20 & 76.92$\pm$2.66 & \textbf{81.15$\pm$3.79} & 77.00$\pm$4.29 & 64.21$\pm$8.31 & 83.60$\pm$2.05 \\
				
				5 &
				& 92.08$\pm$6.19 & 84.55$\pm$1.23 & 77.18$\pm$1.99 & 80.64$\pm$3.11 & 77.33$\pm$3.87 & 65.38$\pm$14.31 & 83.11$\pm$1.82 \\
				\midrule								
				
				\textbf{Ours} &
				& \textbf{93.16$\pm$6.10} & \textbf{84.87$\pm$1.68} & \textbf{77.99$\pm$3.16} & 80.99$\pm$2.58 & \textbf{78.00$\pm$4.89} & \textbf{66.54$\pm$7.97} & 83.91$\pm$1.27 \\
				
				\bottomrule
			\end{tabular}
		\end{center}
		\caption{Ablation studies (upper part) and sensitivity analyses (lower part) on biological graphs.}
		\label{table:ablation_bio_c1}
		%\vspace{-0.2in}
	\end{table*}

	\begin{table}
		\footnotesize
		\setlength\tabcolsep{3.5pt}
		\begin{center}
			\begin{tabular}{ @{} l|c|*{4}{c} @{} }
				\toprule
				%\multicolumn{24}{ c }{\bf GTA5$\,\to\,$Cityscapes } \\
				%\midrule
				& & \bf COLLAB & \bf IMDB-B & \bf IMDB-M & \bf RE-B \\
				\midrule
				
				A1 &  
				\multirow{3}{*}{\rotatebox[origin=c]{90}{ablation}} &
				81.44$\pm$1.57 & 64.70$\pm$11.65 & 51.00$\pm$3.10 & 91.45$\pm$2.13 \\
				
				A2 &
				& 82.20$\pm$1.41 & 74.80$\pm$4.18 & 49.80$\pm$6.39 & 92.95$\pm$1.94 \\
				
				A3 &
				& \textbf{83.08$\pm$1.69} & 75.90$\pm$4.36 & 52.00$\pm$1.54 & 92.70$\pm$1.53 \\
				
				%A3b &
				%& 83.06$\pm$2.08 & 76.50$\pm$4.88 & 52.33$\pm$2.38 & 93.15$\pm$1.27 \\
				\midrule
				\textbf{Ours} &
				& 82.86$\pm$1.81 & \textbf{77.20$\pm$4.73} & \textbf{52.80$\pm$2.45} & \textbf{93.15$\pm$1.58} \\
				\midrule
				%\midrule
				
				%CondDisen &  
				%\multirow{2}{*}{\rotatebox[origin=c]{90}{$f_d$}} &
				%82.20$\pm$1.10 & 75.10$\pm$4.91 & 52.27$\pm$1.86 & 93.00$\pm$1.93 \\
				%DynaDisen &
				%& 82.76$\pm$1.75 & 75.30$\pm$2.58 & 51.73$\pm$2.71 & 93.25$\pm$1.21 \\
				
				\midrule				
				2 &  
				\multirow{2}{*}{\rotatebox[origin=c]{90}{$K$}} &
				82.94$\pm$1.66 & 75.60$\pm$6.69 & 52.20$\pm$2.46 & 92.85$\pm$2.12 \\
				
				8 &
				& \textbf{83.10$\pm$1.80} & 74.90$\pm$5.82 & 51.67$\pm$3.99 & \textbf{93.35$\pm$1.73} \\
				
				\midrule
				1 &  
				\multirow{4}{*}{\rotatebox[origin=c]{90}{$R$}} &
				82.52$\pm$1.87 & 74.50$\pm$4.58 & 51.87$\pm$3.23 & 92.45$\pm$1.82 \\
				
				2 &
				& 82.36$\pm$2.03 & 74.80$\pm$4.73 & 51.40$\pm$3.68 & 92.65$\pm$1.73 \\
				
				4 &
				& 83.04$\pm$1.92 & 74.70$\pm$2.87 & 51.67$\pm$3.07 & 93.10$\pm$0.91 \\
				
				5 &
				& 82.64$\pm$1.44 & 74.50$\pm$6.26 & 51.13$\pm$2.93 & 92.85$\pm$1.67 \\												
				
				\midrule
				\textbf{Ours} &
				& 82.86$\pm$1.81 & \textbf{77.20$\pm$4.73} & \textbf{52.80$\pm$2.45} & 93.15$\pm$1.58 \\
				
				\bottomrule
			\end{tabular}
		\end{center}
		\caption{Ablation studies (upper part) and sensitivity analyses (lower part) on social graphs.}
		\label{table:ablation_social_c1}
		%\vspace{-0.1in}
	\end{table}
	
	\subsubsection{Experimental Results}
	We first compare HGCN with existing state-of-the-art graph classification methods on seven biological datasets.
	Our results demonstrate that HGCN achieves the best performance based on the widely used criterion (Table \ref{table:biological_c1}) and competes favorably against three $C^*$- based baselines (Supplementary Table 5).
	In particular, compared with two capsule-based GNNs (i.e., GCAPS-CNN and CapsGNN), we boost of 16.17\%, 6.49\%, 3.37\%, and 2.79\% improvement on ENZYMES, MUTAG, D\&D, and NCI109, respectively.
	The reason is that higher-order statistical moments of local neighbors are used in GCAPS-CNN to build graph capsules, this strategy, however, fails to identify the underlying latent factors which are important in preserving node/graph property and extracting the hierarchical representations.
	Although this limitation can be partially alleviated by using multiple graph channels as reported in CapsGNN, the transformation matrices used in CapsGNN ignore the structure information involved in lower-level capsules.
	In contrast, we disentangle node representations to explicitly consider the entanglement of heterogeneous factors, and propose transformation GNNs to measure structure-aware part-whole relationships.
	%This reveals the benefits of our proposed (i) graph capsules that disentangle node representations by identifying the underlying latent factors, and (ii) transformation GNNs that model part-whole relationships by explicitly considering the structure information of lower-level capsules.
	Furthermore, it should be noted that we achieve the largest performance gains on ENZYMES which has six classes, compared to the rest of datasets with only two classes.
	This observation implies that HGCN is able to capture more complicated and accurate hierarchy for the multiclass classification problem than other methods.
	%To investigate this hypothesis, we visualize the learned routing weights using example graphs from ENZYMES. 
	%As a counterpart, the learned cluster assignments in DIFFPOOL \cite{ying2018hierarchical} are also visualized for comparison, where DIFFPOOL is widely recognized as the benchmark graph pooling model.
	%As shown in Figure \ref{fig:visualization}, HGCN can more accurately identify the graph hierarchy than DIFFPOOL.
	For instance, although assignment matrix-based methods (e.g., DIFFPOOL) are capable of mapping nodes to a set of clusters for structurally simple graphs, its representation power is limited to complex and crowded graphs such as ENZYMES.
	By contrast, our method performs iterative routing to obtain the cluster of agreement, which jointly learns the hierarchical graph representation and provides the necessary deprecation of assignment ambiguity \cite{sabour2017dynamic}.
	
	Table \ref{table:social_c1} and Table 6 (Supplementary) show the performance comparison on four social graph datasets, where the key challenge is to identify strongly connected communities.
	Similarly, we achieve the significant performance improvement over baseline models, indicating that HGCN can better reason about the part-whole relationships in social networks.
	This is also consistent with the fact that highly complex interactions are involved in social graphs, which can be modeled by identifying heterogeneous factors underlying each node.
	Most importantly, considering the entanglement of the latent factors enables more accurate hierarchy learning. 
	
	%\vspace{-3pt}
	\subsubsection{Ablation Studies}
	Comprehensive ablation studies are carried out in this section to understand the contribution of each component (i.e., disentangled graph capsules, capsule layers, and auxiliary graph reconstruction) in our method.
	Specifically, we (i) directly use the input node representation to serve as graph capsules, without considering the entanglement of heterogeneous factors (A1); (ii) remove the residual connection between adjacent capsule layers (A2); and (iii) remove the auxiliary graph reconstruction (A3).
	%; and (iv) use only primary capsules for graph reconstruction (A3b).
	The results illustrated in Table \ref{table:ablation_bio_c1} and Table \ref{table:ablation_social_c1} (upper part) reveal that (i) disentangling node representation allows us to characterize the latent factors underlying each node and in turn more accurately preserve the node/graph properties and capture the part-whole relationship; (ii) combining fine, low layer information with coarse, high layer information provides us an ability to enhance the final graph-level representation; and (iii) graph reconstruction plays an important role in encoding the instantiation parameters of the input graph and enhancing the training stability. 
	%(iv) class capsules are critical to represent the graph-level information. 
	Thus, we reach the conclusion that each component in our method is necessary and contributes to the performance improvement.
	One exception is NCI1 (A2) and COLLAB (A3), where residual connection and auxiliary graph reconstruction bring inferior performance which may be caused by overfitting.
	%The results of criteria $C^*$ can be found in Supplementary.
	
	%mention our advantages again
	
	\subsubsection{Sensitivity Analyses}
	In this section, we analyze the sensitivity of HGCN to the number of latent factors $K=\{2,8\}$ and the number of routing iterations $R=\{1,2,4,5\}$, where our method with setting: $K=4$ and $R=3$. 
	% the disentangle functions $f_d \in \{\text{CondDisen}, \text{DynaDisen}\}$,
	%Inspired by CondConv \cite{yang2019condconv} and DynaConv \cite{chen2020dynamic}, we  propose two conditional disentangle functions, i.e., CondDisen and DynaDisen.  
	%The results show that more complicated disentanglements only have narrow impact on the performance, implying the robustness of our linear disentangle function.
	%Another possible explanation is that the diversity of the graph datasets in our study is limited.
	As shown in Table \ref{table:ablation_bio_c1} and Table \ref{table:ablation_social_c1} (lower part), the results demonstrate that HGCN is not very sensitive to these two hyper-parameters.
	Although $K=8$ brings limited performance improvement on COLLAB and RE-B than $K=4$ (ours), the computational complexity is doubled in calculating the disentangled graph capsules.
	Similarly, $R=4$ requires more routing iterations, albeit with 0.16\% accuracy boost on D\&D.

	\section{Conclusion}
	In this paper, we introduce a novel HGCN framework for graph classification, which is able to explicitly extract hierarchical graph representations.
	Built upon disentangled graph capsules by identifying heterogeneous factors behind each node, HGCN encodes part-whole relationships by considering the structure information of lower-level parts and iteratively infer the pose of higher-level objects.
	Empirical studies demonstrate the superiority of our framework over existing graph classification methods on 11 commonly used benchmarks.
	
	\section{Acknowledgments}
	This work was partially supported by US National Science Foundation IIS-1718853, the CAREER grant IIS-1553687 and Cancer Prevention and Research Institute of Texas (CPRIT) award (RP190107).
	
	\bibliography{references.bib}
	\bibliographystyle{aaai}
	
\end{document}